\RequirePackage{tikz,pgfplots}

\documentclass[sn-aps,pdflatex,iicol]{sn-jnl}

\jyear{2022}%

\usepackage{caption,subcaption}
\usepackage{graphicx}

\newcommand{\transpose}{$^\mathsf{T}$}

\definecolor{gray}{HTML}{555555}
\definecolor{trans}{HTML}{283593}
\definecolor{other}{HTML}{dbd4d3}
\definecolor{industry}{HTML}{db79c0}
\definecolor{water}{HTML}{7986CB}
\definecolor{techinfra}{HTML}{303355}
\definecolor{orangeme}{HTML}{ED553B}
\definecolor{yellowme}{HTML}{F6D55C}
\definecolor{turquoiseme}{HTML}{3CAEA3}
\definecolor{bluelightme}{HTML}{20639B}
\definecolor{bluedarkme}{HTML}{173F5F}

\usepackage{bm}

\begin{document}

\title[On The Potential of Fractals]{On The Potential of The Fractal Geometry and The CNNs' Ability to Encode it  
}


\author[1]{\fnm{Julia} \sur{El Zini}}\email{jwe04@mail.aub.edu}

\author[1]{\fnm{Bassel} \sur{Musharrafieh}}\email{bam17@mail.aub.edu}

\author*[1]{\fnm{Mariette} \sur{Awad}}\email{mariette.awad@aub.edu.lb}

\affil[1]{\orgdiv{Electrical and Computer Engineering Department}, \orgname{American University of Beirut}, \orgaddress{\street{Riad El-Solh}, \city{Beirut}, \postcode{1107 2020}, \country{Lebanon}}}

\abstract{
The fractal dimension provides a statistical index of object complexity by studying how the pattern changes with the measuring scale. Although useful in several classification tasks, the fractal dimension is under-explored in deep learning applications. In this work, we investigate the features that are learned by deep models and we study whether these deep networks are able to encode features as complex and high-level as the fractal dimensions. Specifically, we conduct a correlation analysis experiment to show that deep networks are not able to extract such a feature in none of their layers. We combine our analytical study with a human evaluation to investigate the differences between deep learning networks and models that operate on the fractal feature solely.  
Moreover, we show the effectiveness of fractal features in applications where the object structure is crucial for the classification task. We empirically show that training a shallow network on fractal features achieves performance comparable, even superior in specific cases, to that of deep networks trained on raw data while requiring less computational resources. Fractals improved the accuracy of the classification by 30\% on average while requiring up to 84\% less time to train. We couple our empirical study with a complexity analysis of the computational cost of extracting the proposed fractal features, and we study its limitation. }


\maketitle

\section{Introduction}\label{sec:intro}
The concept of fractal geometry was developed by Mandelbrot \cite{mandelbrot1977fractals} to model complex objects in non-integer dimensions. A significant number of applications have been studying the fractal geometry of objects in many areas. For instance, the fractal dimension (FD) has been successfully applied in soil particle-size distribution to provide a unique quantitative characterization of the soil data spatial distributions and variability \cite{kravchenko1999multifractal}. 
In medicine, FDs have been applied in the identification of brain tumors from magnetic resonance images \cite{iftekharuddin2003fractal}, detection of micro-calcification in mammograms \cite{shanmugavadivu2012fractal}, malignancy of skin lesions \cite{dobrescu2010medical} and lung tumors \cite{mihara1998usefulness}. Fractal analysis has also been applied in industry and production where FDs are used to perform texture and color analysis and pattern classification to detect defective products \cite{xie2008review}. 
In image segmentation and object detection, fractal models are used to describe the natural scenes and detect man-made objects from images based on the fractal difference between artificial targets and natural background \cite{tian2007detection,wang2016man}. In \cite{backes2009plant,florindo2012fractal}, the fractal descriptors are applied to the analysis of the texture images. Numerous face recognition applications relied on the fractal theory. For instance, the FD has been used in pixel-to-pixel face matching \cite{kouzani1999face} and face recognition \cite{blackledge2009texture}. Recently, Joardar et al.  \cite{face2019enhanced} presented an enhanced patch-wise feature extraction technique based on the FD to develop a pose-invariant face recognition system. However, most of the references discussed above are more than 5 years old and do not make a clear comparison to DL techniques.

Motivated by the potential that fractal geometry possesses, we investigate the ability of Convolutional Neural Networks (CNNs) to encode it. This is currently achieved by Explainable AI (ExAI) researchers through the dissection of the inner workings of DL models  \cite{bau2017network,fong2018net2vec,zeiler2014visualizing,mahendran2015understanding,dosovitskiy2016inverting}. Measuring similarities between deep representations aids in the investigation of what DL models are encoding. Canonical Correlation Analysis (CCA) \cite{hardoon2004canonical} has been used to measure similarity between modeled and measured brain activity \cite{sussillo2015neural}, to train word embeddings in multi-lingual language models \cite{faruqui2014improving} and to compare deep representations in an interpretability framework \cite{raghu2017svcca}. The centered Kernel Alignment (CKA) method has been later introduced in \cite{kornblith2019similarity} to identify similarities in general conditions.

In this work, we utilize CCA and CKA analysis to show that DL models are not able to encode the fractal geometry. For this purpose, we propose a method to extract the fractal dimension at different granularity levels and we conduct a correlation analysis to prove that DL models are still not able to encode fractals neither in their early layers where low-level features are usually extracted nor in deeper levels where higher-level abstractions are formed. 

Furthermore, we study the importance of the extracted features in classification tasks where self-substructure is essential. We show that training shallow models on the fractal features achieves a performance comparable to DL, and sometimes even superior in use cases such as agriculture, remote sensing, and industry while requiring less training time and computational resources. We augment our study with a complexity analysis, empirical robustness guarantees, and a human evaluation of the agreement between DL models and models that operate on the fractal feature solely. 

Next, we start by providing a literature review on the network representation dissection methods and the fractal geometry in Section~\ref{sec:lit}. We then describe how the fractal features can be extracted from a digital image to be correlated with the hidden representations of DL models in Section~\ref{sec:metho}. Then, we present the experimental setup and results in Section~\ref{sec:exp} before concluding with final remarks in Section~\ref{sec:conc}.

\section{Related Work}\label{sec:lit}
\subsection{Feature Encoding in Deep Learning Models}

In an attempt to enhance the interpretability of DL models, some researchers investigated the way patterns are encoded in the hidden layers of a deep network. Such methods provide a quantitative interpretation of the latent representations of deep models and study the relevance of specific features/representations in classification \cite{dovsilovic2018explainable}.

\textit{Network dissection} \cite{bau2017network}, for instance, is a framework that provides a quantitative interpretation of the latent representations of CNNs by studying the alignment between CNNs' hidden units and a set of user-defined semantic concepts. 
Layer-wise Relevance Propagation (LRP) \cite{bach2015pixel}, is a well-used method in the exAI community where the back-propagated signal is interpreted as relevant to understanding classification decisions. LRP allows the visualization of the contributions of the single-pixel in kernel-based classifiers as heatmaps that can be interpreted and validated by human experts. 
Other visualization methods include but are not limited to \cite{zeiler2014visualizing,mahendran2015understanding,dosovitskiy2016inverting}. 

On the other hand, \cite{leino2018influence} propose an \textit{influence} measure that is used to explain the \textit{essence} of a particular class in a classification task and to identify generally-influential neurons over a distribution. 
CCA \cite{hardoon2004canonical} has been also used to measure similarity between modeled and measured brain activity \cite{sussillo2015neural}, to train word embeddings in multi-lingual language models \cite{faruqui2014improving} and to compare deep representations in an interpretability framework \cite{raghu2017svcca}. \cite{kornblith2019similarity} argues that CCA cannot measure meaningful similarities when the layer's representation is of a higher dimension than $n$. \cite{kornblith2019similarity} introduce the CKA method closely related to CCA but can further identify similarities between representations in networks trained from different initialization.

A rigorous evaluation of these approaches, besides other explainability methods, has enabled unprecedented breakthroughs in a variety of explainable computer vision and natural language processing applications \cite{tjoa2020survey,li2020survey}. For instance, CNNs are shown to localize objects without being explicitly trained on the localization tasks in \cite{bazzani2016self}. RNNs in general, and LSTMs and transformers specifically, are also shown to encode specific syntactic/semantic features in specific layers \cite{tenney2019bert}.

\subsection{Fractal Geometry}

The concept of fractal geometry was developed by Mandelbrot \cite{mandelbrot1977fractals} to model complex objects in non-integer dimensions. A significant number of applications have been studying the fractal geometry of objects in many areas. For instance, the fractal dimension (FD) has been applied in the detection of micro-calcification in mammograms \cite{shanmugavadivu2012fractal} and malignancy of skin lesions \cite{dobrescu2010medical}. Fractal analysis has been also applied in industry and production to perform texture and pattern classification to detect defective products \cite{xie2008review}.

Moreover, the fractal theory has been widely applied in image segmentation and object detection. In \cite{wang2016man}, fractal models are used to describe natural scenes and detect man-made objects from images based on the fractal difference between artificial targets and natural backgrounds. In \cite{florindo2012fractal}, the fractal descriptors are applied to analyze the images' texture. Recently, Joardar et al. \cite{face2019enhanced} presented an enhanced patch-wise feature extraction technique based on the FD to develop a pose-invariant face recognition system. 

It is worth mentioning that given the breakthroughs in DL in the past few years, the interest in feature extraction in general and fractal features specifically has significantly declined. This explains why most of the discussed work here on fractal geometry is more than five years old. Moreover, the word \textit{``fractal''} has been recently used in the literature to describe a design strategy for the macro-architecture of neural networks \cite{larsson2016fractalnet,zhou2014object,ning2017knowledge}. These networks do not compute the fractal dimension that was proposed in \cite{mandelbrot1977fractals} which is adopted in this work. They, instead, use the term ``fractal'' to refer to the residual architecture of their deep networks.

\section{Methodology}\label{sec:metho}
In this work, we are interested in researching the following question: \textit{are DL models able to encode the fractal geometry without explicit training?} Thus, we consider the learned representations in deep models at different layers and we correlate them to the extracted fractal features using CCA and CKA analyses. We start by detailing how the fractal feature vector is extracted from a digital image. Then, we describe the correlation methods that we follow to study the representation similarity.

\subsection{Fractal Features}
\subsubsection{Fractal Dimension Computation}\label{fractal_image}
The fractal dimension $ FD =  \frac{\ln (N)}{\ln (1/r)} \in \mathbb{R}$ of an object is defined as the exponent of the number of self-similar pieces, $N$, with a magnification factor, $1/r$, into which a figure may be shattered. In a digital image, $FD$ is estimated using the box-counting method \cite{sarkar1992efficient} which estimates the number of boxes with side length $r$ that are needed to cover the surface of a fractal object and the number of grid boxes $N$ occupied by one or more pixels of the digital image. Once $N$ and $r$ are found, FD is estimated as the slope of the line that best fits the 2D points of the form $(\ln(N), \ln(1/r))$. 

\subsubsection{From Fractal Dimension to Fractal Features}\label{sec:sub}
As defined, the fractal dimension is a scalar that reflects the self-structure in an image. We define the fractal feature vector to be an integration of the fractal dimension of different sub-images at different granularity levels. For this purpose, we consider non-overlapping sub-regions of the image of different sizes $w\times w$, and we compute their fractal dimensions. To provide robustness to scaling and shifting, we consider different window sizes $w$, merging all results into one feature vector. 

\begin{figure}[h]
	\centering
	\includegraphics[width=0.45\textwidth]{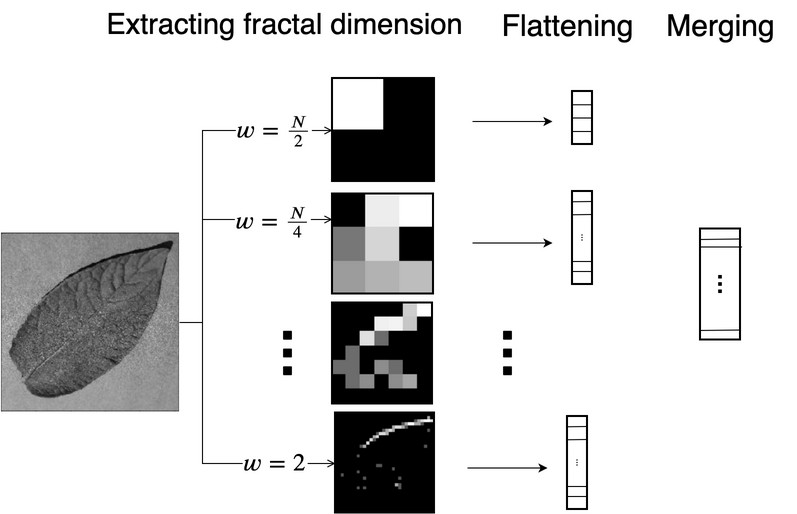}
	\caption{Computing the fractal features of an $M\times M$ image}\label{fig:metho}
\end{figure}

Figure~\ref{fig:metho} illustrates how the fractal features are computed for a $M\times M$ image by changing the window size from $w=2$ to $w=\frac{M}{2}$, computing the fractal feature under each $w\times w$ block then flattening and merging all the results. This formulation enables the study of an object at different granularity levels by tracking how its internal substructure evolves at different \textit{zoom in} levels. Once the vector of the zoomed-in fractal features (ZFrac) is computed, it is fed into a shallow feed-forward Neural Network (ZFrac+NN) to perform the classification instead of feeding the raw image. 

\subsection{Representation Similarity}
Formally, a layer representation is the mapping between all possible inputs and outputs. Since infinite mappings are not practically feasible, we consider the layer's representation to be its response over a finite set of inputs drawn from a particular dataset. Specifically, for a given dataset $X=\{\bm{x}_1, \dots, \bm{x}_n\}$ and a layer $l$, the layer's representation is defined by $\bm{Y} \in \mathbb{R}^{n\times p_2} = \big[ \bm{z_l}(\bm{x_1})$\transpose$, \dots, \bm{z_l}(\bm{x_n})$\transpose$ \big]$\transpose
with $\bm{z}_l(\bm{x}_i) \in \mathbb{R}^{p_2}$ is the concatenation of all the neurons' output in $l$ on the input $x_i$. Let $Z\in\mathbb{R}^{n\times p_1}$ denote a matrix of fractal features ZFrac for the same $n$ examples.

\subsubsection{CCA Analysis}
Once the fractal features $\bm{Z}$ and the layers representations $\bm{Y}$ are extracted, CCA method linearly transforms $\bm{Z}$ and $\bm{Y}$ to new sub-spaces $\bm{\Tilde{Z}} = \bm{W_1}\bm{Z}$ and $\bm{\Tilde{Y}} = \bm{W_2}\bm{Y}l$ to achieve maximum alignment and then computes their correlation score.

\subsubsection{CKA Analysis}
CKA is a normalized index derived from the Hilbert-Schmidt Independence Criterion, HSIC, which is computed as follows: 
\begin{equation}
    CKS(Z,Y) = \frac{HSIC(\bm{Z}\bm{Y})}{\sqrt{HSIC(\bm{Y}\bm{Y}) .  HSIC(\bm{Z}\bm{Z})}}
\end{equation}
We consider linear and RBF kernels. Based on the recommendation of the authors in \cite{kornblith2019similarity}, we set $\sigma$ in the RBF kernel to be a fraction $\alpha$ of the median distance between the $n$ examples. 

CCA and CKA scores are computed across all layers in a DL model. We seek peaks in the correlation score to localize the layer that is responsible for the extraction of the feature in question. The absence of peaks suggests the inability of the network to encode the corresponding feature.

\section{Experiments}\label{sec:exp}
We first show the inability of DL models to encode the fractal feature through the CCA and the CKA experiments. We complement our study with a human evaluation to study whether fractal-based and DL models differ for a human evaluator. Finally, we show the potential of fractal-based networks in specific classification tasks. 

\subsection{Experimental Setup}
For this experiment, an Intel(R) Core(TM) i7 machine with a 4-core CPU at 2.60GHz and 2601 Mhz is considered with \textit{Python 3.5 } and \textit{Keras} backend. ZFrac+NN consists of 2 fully connected layers with 100 and 50 neurons respectively and a \textit{relu} activation. Training ZFrac+NN and DL models is done using \textit{Adam} optimizer for 200 epochs with an early stopping activated after each 3-epoch plateau for the validation loss. 

\begin{figure*}[h]
    \centering
    \includegraphics[width=0.9\textwidth]{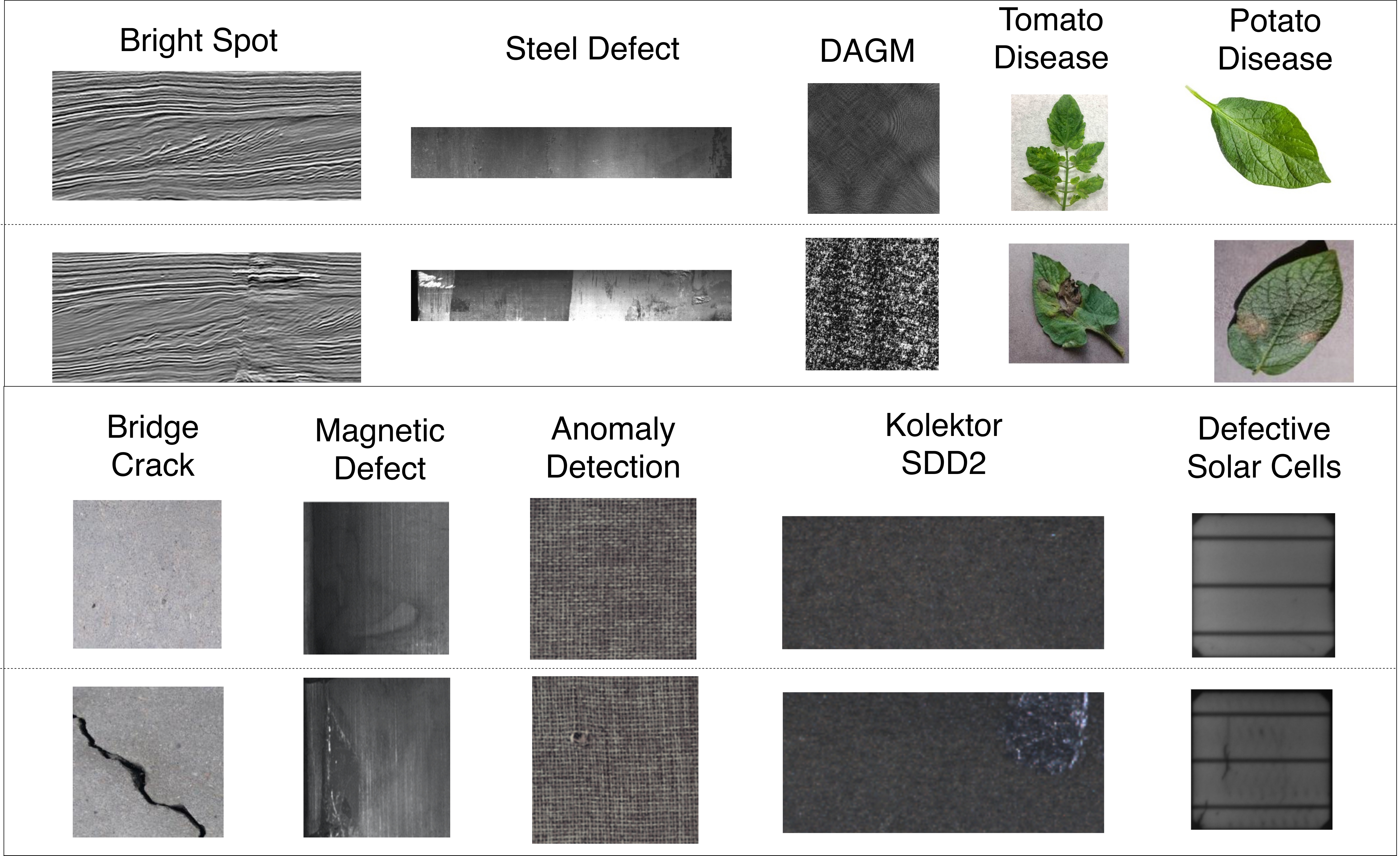}
    \caption{Samples from the datasets considered in this work. For binary classification datasets, we illustrate one random image from each category. For multi-class classification datasets, we choose two categories at random and we visualize one random image from each. \textit{(The images here are resized for display purposes)}}
    \label{fig:data_samples}
\end{figure*}

We consider 8 different datasets with characteristics summarized in Table~\ref{tbl:datasets}. Self-structure is essential for the success of the classification task in the datasets considered in this work. For instance, the purpose of the bright spot detection dataset \cite{el2019deep} is to identify seismic attribute anomalies that can indicate the presence of hydrocarbon formation. The potato and tomato disease datasets \cite{hughes2015open} require recognizing unusual structures in leaves that implied a plant disease. Similarly, the detection of natural and industrial defects is needed in the DAGM,  steel defect detection, bridge crack detection \cite{xu2019automatic} datasets, magnetic tile defect \cite{huang2020surface}, anomaly detection \cite{bergmann2019mvtec} Electroluminescence defective solar cells\cite{buerhop2018benchmark} and Kolektor \cite{bovzivc2021mixed} datasets. Figure~\ref{fig:data_samples} shows two sample images from each dataset considered in this work. 

	\begin{table*}[h]
			\small\centering
		\begin{tabular}{lrrrr}
			\textbf{\small{Dataset}} & \textbf{\small{Classes}} & \textbf{\small{\# train}} & \textbf{\small{\# test}}  &\textbf{\small{Image size}}\\
			\hline\hline
			\small{Bright Spot } & 2 & 70 & 30 & 4,017x1,690\\
			\small{Steel Defect\footnote{\url{kaggle.com/c/severstal-steel-defect-detection }}}  & 4 & 3,500 & 1,500& 1,600x256 \\
			\small{DAGM} \footnote{\url{resources.mpi-inf.mpg.de/conference/dagm/2007/prizes.html}}& 6 & 1,440 & 360 & 512x512\\
			\small{Tomato disease} & 2& 2,780& 720 & 256x256\\
			\small{Potato disease} & 2& 806& 346& 256x256\ \\
			Bridge Crack & 2 & 4,856 & 1213 & 224x224\\
			Magnetic Defect & 6 & 1,075& 269 & 224x224 \\
			Anomaly Detection & 2 & 3,629 & 1,725 & 1,024x1,024\\
			Kolektor SDD2 & 2 & 2,331 & 1,004 & 230x630\\
			Electroluminescence  & & & & \\
			Defective Solar Cells & 2 & 2,099 & 525 & 300x300\\
			\hline
			\hline
		\end{tabular}
	\caption{Experimental datasets description}\label{tbl:datasets}
	\end{table*}

\subsection{Representation Similarity}
\subsubsection{Correlation Results}
We consider $150$ randomly chosen samples from the steel defect detection \textit{testing} dataset. These samples are then fed to four state-of-the-art (SOTA) deep models, \textit{VGG19} \cite{vgg}, \textit{InceptionV3} \cite{inception}, \textit{ResNet152V2} \cite{he2016deep}, and \textit{DenseNet201} \cite{densenet}. In each network, all layer's activations are recorded on the $150$ samples that are correlated with the extracted fractal features at different granularity levels ($2,4,8,16$ and $32$ specifically). The maximum CCA and CKA scores across all layers are reported in Table~\ref{tbl:cca_fractals}. We reinforce our study by computing three other correlation metrics (Pearson, Spearman, and Kendall) and by reporting the maximum in Table~\ref{tbl:cca_fractals}. Furthermore, we compare the correlation between deep representations and the fractal features to that between deep representations and other low-level features that are widely adopted in the computer vision community, namely: the Prewitt edges (horizontal and vertical), Harris corners, Histogram of Oriented Gradients (HoG), and GLCM texture features. 

\begin{table*}[h]
\centering
\small
\begin{tabular}{lllllllllll}
\hline
Metric     & \rotatebox{90}{CKA}               & \rotatebox{90}{CCA}                & \rotatebox{90}{Pearson} & \rotatebox{90}{Spearman} & \rotatebox{90}{Kendall} & \rotatebox{90}{CKA} & \rotatebox{90}{CCA} & \rotatebox{90}{Pearson} & \rotatebox{90}{Spearman} & \rotatebox{90}{Kendall} \\
\hline
Feature    & \multicolumn{5}{c}{VGG19}       & \multicolumn{5}{c}{Inception V3}  \\
\cmidrule(lr){2-6} \cmidrule(lr){7-11}

Fractal    & 0.1                                                  & 0.08                                                  & 0.02                                       & 0.05                                        & 0.04                                       & 0.09                                   & 0.06                                   & 0.04                                       & 0.04                                        & 0.01                                       \\
GLCM       & \textbf{1.00}                                        & \textbf{1.00}                                         & 0.87                                       & \textbf{0.98}              & 0.92                                       & \textbf{1.00}         & \textbf{1.00}         & 0.93                                       & 0.76                                        & 0.82                                       \\
Harris     & 0.85                                                 & 0.71                                                  & 0.62                                       & 0.54                                        & 0.74                                       & 0.62                                   & 0.65                                   & 0.66                                       & 0.65                                        & 0.53                                       \\
HoG        & \textbf{1.00}                       & \textbf{1.00}                        & 0.93                                       & 0.82                                        & 0.82                                       & \textbf{1.00}         & \textbf{1.00}         & \textbf{0.97}             & \textbf{1.00}              & 0.98                                       \\
Prewitt\_H & 0.89                                                 & 0.51                                                  & 0.57                                       & 0.61                                        & 0.68                                       & 0.85                                   & 0.45                                   & 0.42                                       & 0.46                                        & 0.62                                       \\
Prewitt\_V & 0.87                                                 & 0.99                                                  & \textbf{0.97}             & 0.89                                        & \textbf{0.97}             & 0.81                                   & 0.91                                   & \textbf{0.97}             & 0.99                                        & \textbf{1.00}             \\
\hline
Feature    & \multicolumn{5}{c}{ResNet152V1} & \multicolumn{5}{c}{DenseNet201}  \\
\cmidrule(lr){2-6} \cmidrule(lr){7-11}
Fractal    & 0.06                                                 & 0.04                                                  & 0.1                                        & 0.09                                        & 0.1                                        & 0.09                                   & 0.06                                   & 0.03                                       & 0.07                                        & 0.09                                       \\
GLCM       & \textbf{0.99}                       & 0.67                                                  & 0.63                                       & 0.32                                        & 0.59                                       & 0.98                                   & 0.84                                   & 0.82                                       & \textbf{1.00}              & \textbf{0.97}             \\
Harris     & 0.67                                                 & 0.62                                                  & 0.58                                       & 0.59                                        & 0.61                                       & 0.72                                   & 0.63                                   & 0.65                                       & 0.73                                        & 0.71                                       \\
HoG        & 0.87                                                 & 0.51                                                  & 0.63                                       & 0.53                                        & 0.39                                       & \textbf{1.00}         & \textbf{0.99}         & \textbf{0.93 }            & \textbf{1.00}              & 0.93                                       \\
Prewitt\_H & 0.89                                                 & \textbf{0.73}                        & \textbf{0.72}             & 0.71                                        & \textbf{0.72}             & 0.91                                   & 0.53                                   & 0.57                                       & 0.45                                        & 0.32                                       \\
Prewitt\_V & 0.93                                                 & 0.59                                                  & 0.64                                       & 0.82                                        & 0.53                                       & 0.96                                   & \textbf{0.82}         & 0.78                                       & 0.93                                        & 0.8  \\
\hline\hline
\end{tabular}
\caption{Maximum correlation between the corresponding feature and the hidden representations of the four DL models. Highest correlations are highlighted in bold. }\label{tbl:cca_fractals}
\end{table*}

As reported in Table~\ref{tbl:cca_fractals}, the CKA and CCA correlation between the extracted fractal features and the layer representations was found to be consistently low ($\le 0.1$) across all DL models, layers, and correlation metrics. This low correlation suggests that the DL models that we considered are lacking the ability to encode the fractal features. Experiments on bright spot detection, DAGM, and Tomato and Potato disease, show similar results and are reported in the appendix. The results also reveal that the maximum correlation achieved by the fractal features is consistently lower than the other low-level features. The latter can reach a high correlation score (up to 100\%) which suggests an obvious encoding of the corresponding feature in a particular layer. Finally, as expected, CKA coefficients are consistently higher than those of CCA. This shows that the CKA method is able to measure more meaningful similarities between representations as suggested in \cite{kornblith2019similarity}. 

\subsubsection{Correlation Across Layers}
\begin{figure}[h]
    \centering
    \begin{subfigure}[h]{0.24\textwidth}
      \centering
      \includegraphics[width=0.9\textwidth]{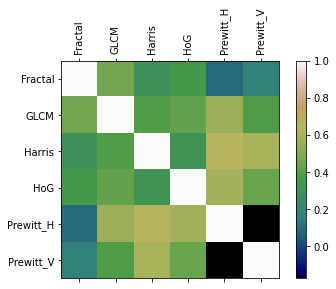}
      \caption{CKA}
    \end{subfigure}%
    \begin{subfigure}[h]{0.24\textwidth}
      \centering
      \includegraphics[width=0.9\textwidth]{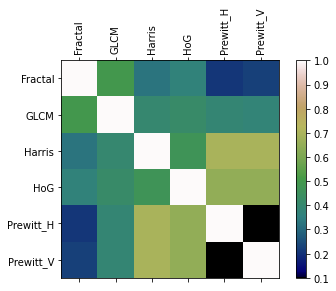}
      \caption{CCA}
    \end{subfigure}
    \caption{Maximum CKA and CCA correlation score between every pair of features averaged across all layers on DenseNet201 and the steel defect detection data}\label{fig:cka_baseline}
\end{figure}

In order to visualize how correlation scores change when the depth of the network is varied, we plot the CCA scores across different layers for the fractal feature and the low-level features in the four DL models. The results in Figure~\ref{fig:cca_graph} demonstrate that the fractal feature maintains a consistently negligible correlation with the deep representations of the deep networks across different layers in comparison to other low-level features according to their CCA and CKA scores. The graphs also suggest that none of the layers of the considered deep networks is attempting at extracting the fractal features. This can be explained by the pre-eminent pattern of the other low-level features that exhibit a correlation evident in the clear peaks that are missing in the fractal correlation graph. As a baseline, we visualize the maximum CCA and CKA between every pair of features in Figure~\ref{fig:cka_baseline}. The correlations show a relatively low association between fractals and the other low-level features (as compared to the edges (Prewitt) and oriented gradients (HoG)). 


\pgfplotsset{compat = 1.10}
\pgfplotstableread[col sep = comma]{vgg_cka.csv}\datavggcka

\pgfplotstableread[col sep = comma]{inception_cka.csv}\datainceptioncka

\pgfplotstableread[col sep = comma]{resnet_cka.csv}\dataresnetcka

\pgfplotstableread[col sep = comma]{densenet_cka.csv}\datadensenetcka

\pgfplotsset{compat = 1.10}
\pgfplotstableread[col sep = comma]{vgg.csv}\datavgg

\pgfplotstableread[col sep = comma]{inception.csv}\datainception

\pgfplotstableread[col sep = comma]{resnet.csv}\dataresnet

\pgfplotstableread[col sep = comma]{densenet.csv}\datadensenet

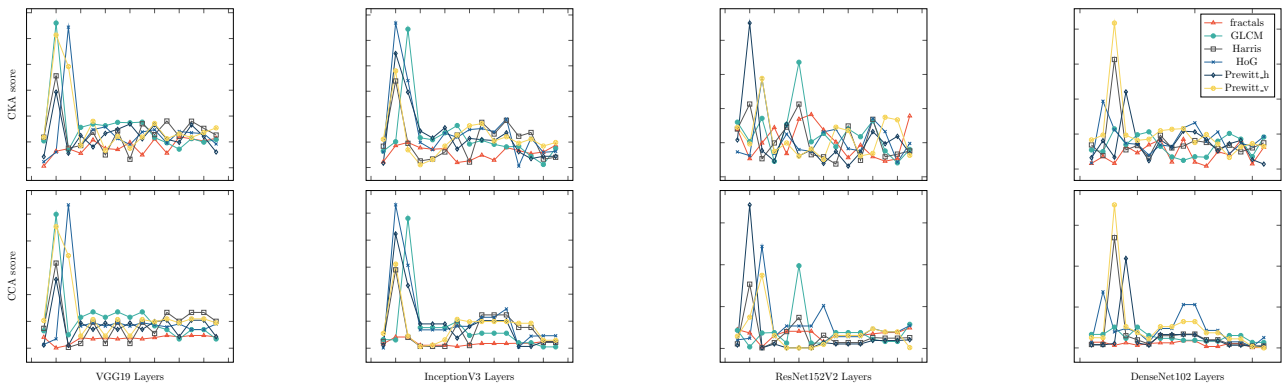
\begin{figure*}[h]
\centering
\begin{subfigure}[b]{0.12\textwidth}
    \begin{tikzpicture}[scale=.4]
      \begin{axis}[
    legend pos= north east,
    yticklabels={,,},
    xticklabels={,,},
    ylabel={CKA score},
    ]
        \addplot[orangeme,mark=triangle] table[x index = {0}, y index = {1}]{\datavggcka};
        \addplot[turquoiseme,mark=*] table[x index = {0}, y index = {2}]{\datavggcka};
        \addplot[gray,mark=square] table[x index = {0}, y index = {3}]{\datavggcka};
        \addplot[bluelightme,mark=x]  table[x index = {0}, y index = {4}]{\datavggcka};
        \addplot[bluedarkme,mark=diamond]  table[x index = {0}, y index = {5}]{\datavggcka};
        \addplot[yellowme,mark=otimes]  table[x index = {0}, y index = {6}]{\datavggcka};
      \end{axis}
    \end{tikzpicture}
    \end{subfigure}
        \hfill
    \begin{subfigure}[b]{0.12\textwidth}
\begin{tikzpicture}[scale=.4]
  \begin{axis}[
    legend style={at={(0.5,-0.15), font=\footnotesize},
    anchor=north,legend columns=-1},
    yticklabels={,,},
    xticklabels={,,},
    ]
    \addplot[orangeme,mark=triangle] table[x index = {0}, y index = {1}]{\datainceptioncka};
    \addplot[turquoiseme,mark=*] table[x index = {0}, y index = {2}]{\datainceptioncka};
    \addplot[gray,mark=square] table[x index = {0}, y index = {3}]{\datainceptioncka};
    \addplot[bluelightme,mark=x]  table[x index = {0}, y index = {4}]{\datainceptioncka};
    \addplot[bluedarkme,mark=diamond]  table[x index = {0}, y index = {5}]{\datainceptioncka};
    \addplot[yellowme,mark=otimes]  table[x index = {0}, y index = {6}]{\datainceptioncka};
  \end{axis}
\end{tikzpicture}
\end{subfigure}
\hfill
     \begin{subfigure}[b]{0.12\textwidth}
\begin{tikzpicture}[scale=.4]
  \begin{axis}[
    legend style={at={(0.5,-0.15), font=\footnotesize},
    anchor=north,legend columns=-1},
    yticklabels={,,},
    xticklabels={,,},
    ]
    \addplot[orangeme,mark=triangle] table[x index = {0}, y index = {1}]{\dataresnetcka};
    \addplot[turquoiseme,mark=*] table[x index = {0}, y index = {2}]{\dataresnetcka};
    \addplot[gray,mark=square] table[x index = {0}, y index = {3}]{\dataresnetcka};
    \addplot[bluelightme,mark=x]  table[x index = {0}, y index = {4}]{\dataresnetcka};
    \addplot[bluedarkme,mark=diamond]  table[x index = {0}, y index = {5}]{\dataresnetcka};
    \addplot[yellowme,mark=otimes]  table[x index = {0}, y index = {6}]{\dataresnetcka};
  \end{axis}
\end{tikzpicture}
\end{subfigure}
\hfill
     \begin{subfigure}[b]{0.12\textwidth}
\begin{tikzpicture}[scale=.4]
  \begin{axis}[
  legend pos= north east,
    yticklabels={,,},
    xticklabels={,,},
    ]
    \addplot[orangeme,mark=triangle] table[x index = {0}, y index = {1}]{\datadensenetcka};
    \addplot[turquoiseme,mark=*] table[x index = {0}, y index = {2}]{\datadensenetcka};
    \addplot[gray,mark=square] table[x index = {0}, y index = {3}]{\datadensenetcka};
    \addplot[bluelightme,mark=x]  table[x index = {0}, y index = {4}]{\datadensenetcka};
    \addplot[bluedarkme,mark=diamond]  table[x index = {0}, y index = {5}]{\datadensenetcka};
    \addplot[yellowme,mark=otimes]  table[x index = {0}, y index = {6}]{\datadensenetcka};
    \legend{fractals,GLCM,Harris,HoG,Prewitt\_h,Prewitt\_v}
  \end{axis}
\end{tikzpicture}
\end{subfigure}


\begin{subfigure}[b]{0.12\textwidth}
    \begin{tikzpicture}[scale=.4]
      \begin{axis}[
    legend pos= north east,
    yticklabels={,,},
    xticklabels={,,},
    xlabel={VGG19 Layers},
    ylabel={CCA score},
    ]
        \addplot[orangeme,mark=triangle] table[x index = {0}, y index = {1}]{\datavgg};
        \addplot[turquoiseme,mark=*] table[x index = {0}, y index = {2}]{\datavgg};
        \addplot[gray,mark=square] table[x index = {0}, y index = {3}]{\datavgg};
        \addplot[bluelightme,mark=x]  table[x index = {0}, y index = {4}]{\datavgg};
        \addplot[bluedarkme,mark=diamond]  table[x index = {0}, y index = {5}]{\datavgg};
        \addplot[yellowme,mark=otimes]  table[x index = {0}, y index = {6}]{\datavgg};
      \end{axis}
    \end{tikzpicture}
    \end{subfigure}
        \hfill
    \begin{subfigure}[b]{0.12\textwidth}
\begin{tikzpicture}[scale=.4]
  \begin{axis}[
    legend style={at={(0.5,-0.15), font=\footnotesize},
    anchor=north,legend columns=-1},
    yticklabels={,,},
    xticklabels={,,},
    xlabel={InceptionV3 Layers},
    ]
    \addplot[orangeme,mark=triangle] table[x index = {0}, y index = {1}]{\datainception};
    \addplot[turquoiseme,mark=*] table[x index = {0}, y index = {2}]{\datainception};
    \addplot[gray,mark=square] table[x index = {0}, y index = {3}]{\datainception};
    \addplot[bluelightme,mark=x]  table[x index = {0}, y index = {4}]{\datainception};
    \addplot[bluedarkme,mark=diamond]  table[x index = {0}, y index = {5}]{\datainception};
    \addplot[yellowme,mark=otimes]  table[x index = {0}, y index = {6}]{\datainception};
  \end{axis}
\end{tikzpicture}
\end{subfigure}
\hfill
     \begin{subfigure}[b]{0.12\textwidth}
\begin{tikzpicture}[scale=.4]
  \begin{axis}[
    legend style={at={(0.5,-0.15), font=\footnotesize},
    anchor=north,legend columns=-1},
    yticklabels={,,},
    xticklabels={,,},
    xlabel={ResNet152V2 Layers},
    ]
    \addplot[orangeme,mark=triangle] table[x index = {0}, y index = {1}]{\dataresnet};
    \addplot[turquoiseme,mark=*] table[x index = {0}, y index = {2}]{\dataresnet};
    \addplot[gray,mark=square] table[x index = {0}, y index = {3}]{\dataresnet};
    \addplot[bluelightme,mark=x]  table[x index = {0}, y index = {4}]{\dataresnet};
    \addplot[bluedarkme,mark=diamond]  table[x index = {0}, y index = {5}]{\dataresnet};
    \addplot[yellowme,mark=otimes]  table[x index = {0}, y index = {6}]{\dataresnet};
  \end{axis}
\end{tikzpicture}
\end{subfigure}
\hfill
     \begin{subfigure}[b]{0.12\textwidth}
\begin{tikzpicture}[scale=.4]
  \begin{axis}[
  legend pos= north east,
    yticklabels={,,},
    xticklabels={,,},
    xlabel={DenseNet102 Layers},
    ]
    \addplot[orangeme,mark=triangle] table[x index = {0}, y index = {1}]{\datadensenet};
    \addplot[turquoiseme,mark=*] table[x index = {0}, y index = {2}]{\datadensenet};
    \addplot[gray,mark=square] table[x index = {0}, y index = {3}]{\datadensenet};
    \addplot[bluelightme,mark=x]  table[x index = {0}, y index = {4}]{\datadensenet};
    \addplot[bluedarkme,mark=diamond]  table[x index = {0}, y index = {5}]{\datadensenet};
    \addplot[yellowme,mark=otimes]  table[x index = {0}, y index = {6}]{\datadensenet};
  \end{axis}
\end{tikzpicture}

\end{subfigure}
\caption{CKA and CCA scores on different layers in the DL models show consistently low correlation with the fractal features in comparison to other low-level features suggesting that such networks are unable to encode the fractal features.}\label{fig:cca_graph}
\end{figure*}

\subsubsection{Agreement}
\begin{figure}[h]
    \centering
    \includegraphics[width=0.35\textwidth]{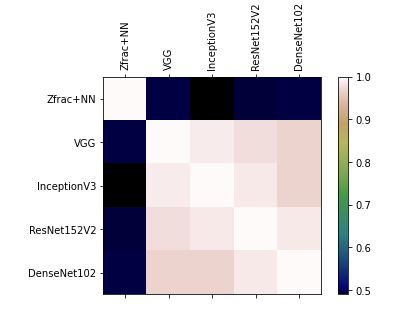}
    \caption{Percentage of agreement on new testing images between ZFrac+NN and the DL models}
    \label{fig:agreement}
\end{figure}

To further investigate the agreement between the fractal dimension and the DL hidden representations we consider ZFrac+NN and we compare its predictions to the DL models on the steel defect detection dataset by studying the agreement percentage between the models. Figure~\ref{fig:agreement} shows that the agreement between ZFrac+NN and the DL models does not exceed 50\% of the test cases while the four DL models agree on almost all the test cases. One can conclude that, even with different architectures, the predictions in DL models are almost consistent; the right, as well as the wrong predictions, agree between different models. Networks based on the fractal feature however can perceive images differently; ZFrac+NN can correctly detect some of the defects that DL models missed and vice versa. This result paves the way to significant improvement on both approaches when \textit{ensembling} methods are exploited.

\subsection{Difference between Z-Frac and DL for a human evaluator}
\begin{figure}[h]
\centering
\begin{subfigure}[h]{0.22\textwidth}
  \centering
  \includegraphics[width=0.9\textwidth]{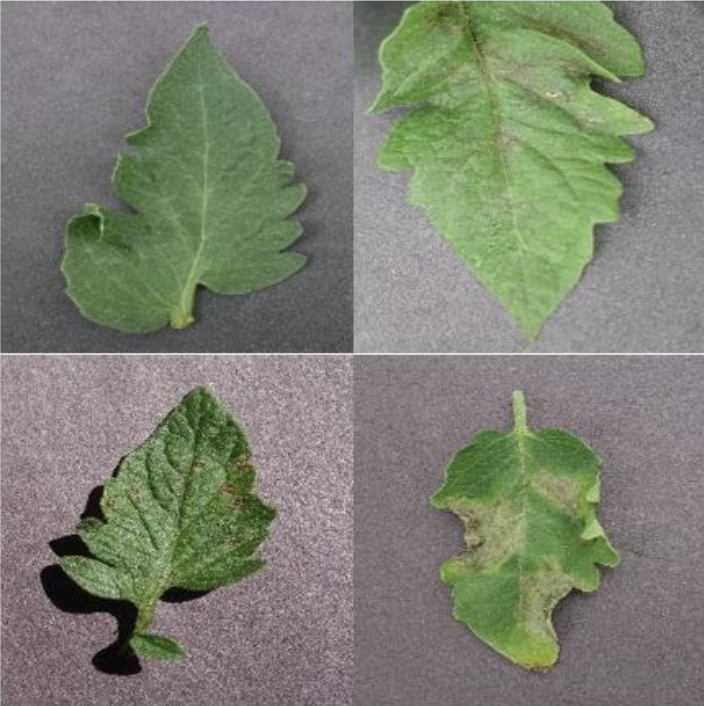}
  \caption{ZFrac+NN}
\end{subfigure}%
\begin{subfigure}[h]{0.22\textwidth}
  \centering
  \includegraphics[width=0.9\textwidth]{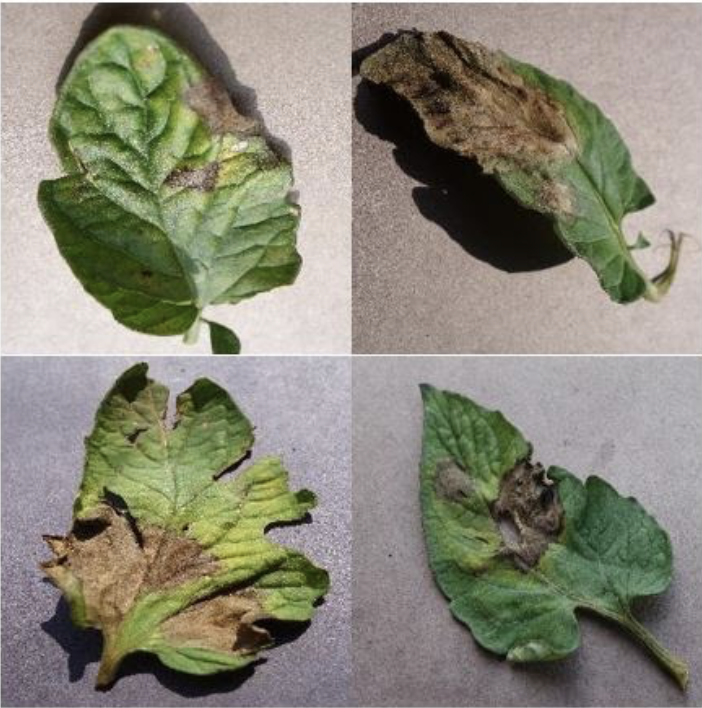}
  \caption{DL}
\end{subfigure}
\caption{Sample wrong predictions from the potato disease dataset where 96.5\% of the participants reported that the disease in (b) is more obvious to detect. }
\label{fig:survey_sample}
\end{figure}

\begin{figure}[h]
\centering
     \begin{subfigure}[b]{0.35\textwidth}
        \includegraphics[width=\textwidth]{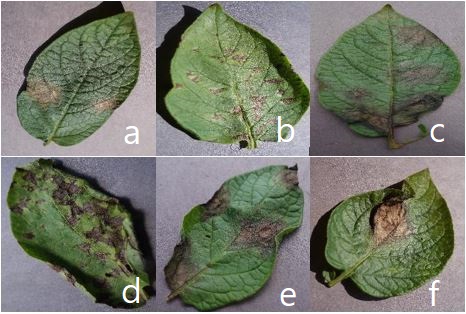}
    \end{subfigure}
    \begin{subfigure}[b]{0.4\textwidth}
            \includegraphics[width=\textwidth]{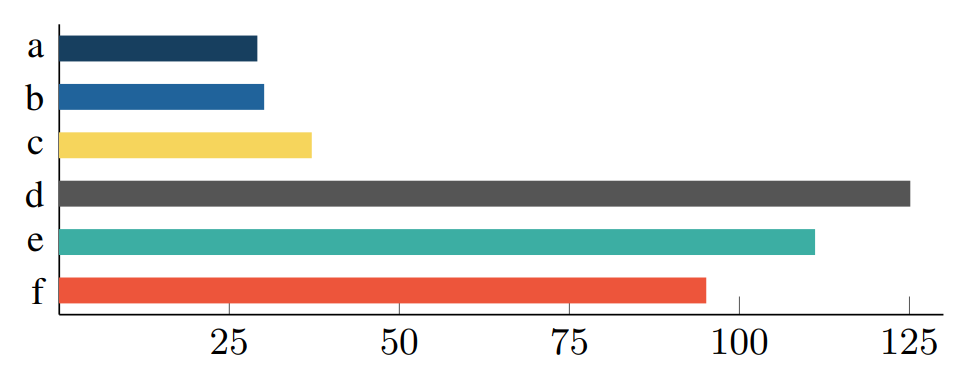}
    \end{subfigure}
    \caption{(Top) Sample wrong predictions by ZFrac+NN (a, b and c) and DL models (d, e and f) where participants are asked to choose the top-3 images where the leaf disease is obvious to detect. (Bottom) Number of times each image is selected as having an obvious leaf disease.  Top-3 images belong to wrong predictions in DL models (d, e, and f).}
    \label{fig:survey_rank}
\end{figure}

We finally conduct a survey on the wrong predictions in ZFrac+NN and DL models. 140 responses were collected from an audience of different age groups.\footnote{$48\%$ of the participants were female and 60\% of the participants had a prior experience in image classification applications} The participants were asked to evaluate the difficulty of some misclassified images that were chosen randomly from the experiments we ran on the bright spot detection, steel defect detection, potato and tomato diseases, and DAGM datasets. Figure~\ref{fig:survey_sample} shows a sample of the wrong predictions that were included in the survey.  72\%, 88\%, 92\% and  96\% and of the participants declared that wrong predictions in the ZFrac+NN are less obvious than those of the DL models on the DAGM, potato, steel defect detection, and tomato disease datasets respectively. When shown 6 misclassified images and asked to choose the hardest 3 images to classify in each dataset as in Figure~\ref{fig:survey_rank}, participants consistently selected the top-1 image to belong to the ZFrac+NN model in all datasets, and on average 2 out of the top-3 images belong to the ZFrac+NN model. The top-3 images selected by the participants to have a disease more obvious to detect belonged all to the DL wrong predictions in the tomato and potato disease dataset as shown in Figure~\ref{fig:survey_rank}. 

These results emphasize the role the fractal features can play in reflecting the nature of objects given that ZFrac+NN misses are less obvious than those missed by DL models. Finally, for the bright spot detection dataset where ZFrac+NN does not fail on the test set, we asked the participants to rank the difficulty of the images where DL models failed. 76\% and 31\% considered the images as easy and extremely easy respectively which highlights the fact that DL models can fail badly when limited training data is available even if the classification task is relatively not challenging. 

\subsection{Potential of ZFrac}
\subsubsection{Performance}
One could argue that fractal features are not encoded in the presented SOTA deep networks because they are irrelevant to the classification task. Although the literature proved the importance of this feature in different classification tasks, we further compare ZFrac+NN trained on the fractal features, solely, to the DL models that were pre-trained on ImageNet and fine-tuned on the raw images. We call the reader's attention that the goal is not to show that training on the fractal features solely surpasses training a deep network on the raw image. We rather aim at highlighting the potential these features possess for yielding performance comparable to the deep models that are pre-trained on millions of images. 

Table~\ref{tbl:performance} presents the results of training ZFrac+NN and DL models on the raw images for different datasets. For the bright spot detection dataset, where limited data is available, ZFrac+NN perfectly predicts bright spots while the DL performance is very poor, probably because the model is only trained on 70 quite high-dimensional images. 
On the steel defect detection, potato and tomato disease, and DAGM datasets where the training data is larger, ZFrac+NN achieves good results. More specifically, it achieves a performance close to that of DL in steel defect detection and improves accuracy by up to 47\% in the tomato and potato disease dataset. For the DAGM data, where a multi-class classification is considered, ZFrac+NN does not achieve the same accuracy as DL models but consistently takes less time to train by up to a 5x speedup factor. 

\begin{table*}[hbt!]
\centering
\small
\begin{tabular}{llrrrr}
\textbf{Dataset}      & \textbf{Model}                    & \textbf{Acc.} & \textbf{F1} & \textbf{Training} & \textbf{Avg. Inf.}\\
\hline\hline
Bright  Spot     & ZFrac+NN           & \textbf{100}                 & \textbf{100}  & \textbf{106}  & \textbf{0.143}                \\
Detection       & VGG19                      & 57       & 53                    & 4,102         &     0.316                     \\
            & InceptionV3                & 61         & 64                  & 5,344            &     0.273                 \\
             & ResNet152V2                   & 62      & 65                     & 5,904          &       0.304                  \\
             & DenseNet102                 & 63           &61                & 5,827             &           0.315           \\
             \hline
Steel Defect  & ZFrac+NN                  & \textbf{74}          & \textbf{71}                 & 7,507     &     \textbf{0.203}                        \\
            & VGG19                      & 71    & 69                      & 8,639        & 0.368                        \\
             & InceptionV3                & \textbf{74}       & 70                 & \textbf{1,156}   &     0.231                         \\
             & ResNet152V2          & 67          & 64               & 3,570                      &   0.371        \\
             & DenseNet102                 & 54        & 55                 & 2,016                &    0.324             \\
             \hline

Potato Disease      & ZFrac+NN           & \textbf{88}            & \textbf{91}               & \textbf{70}     & \textbf{0.014}                                 \\
             & VGG19                      & 54           & 53              & 652                 & 0.392                     \\
             & InceptionV3                & 55            & 47             & 186            &     0.272                      \\
             & ResNet152V2                   & 53          & 42              & 525           &        0.369                   \\
             & DenseNet102                 & 50            & 48               & 378          &         0.307                  \\
             \hline
Tomato Disease      & ZFrac+NN         & \textbf{99}            & \textbf{98}               & 851                      & \textbf{0.004}               \\
             & VGG19                      & 48           & 53           & 1,750                 &  0.312            \\
             & InceptionV3                & 52           &  58             & 682                &    0.224                  \\
             & ResNet152V2                   & 50           & 49                & \textbf{530}          &    0.321                     \\
             & DenseNet102                 & 46            & 47            & 1,008                 &      0.315            \\
             \hline
DAGM         & ZFrac+NN                    & 85            & 82               & \textbf{1,699}              & \textbf{0.007}                     \\
             & VGG19             & \textbf{97}             & \textbf{97}             & 9,607                   & 0.332               \\
             & InceptionV3               & 91              & 92           & 1,864                    &  0.295              \\
             & ResNet152V2                   & 91           & 93              & 6,430                   &   0.351          \\
             & DenseNet102                 & 91            & 92              & 3,841                    &    0.324       \\
             \hline   
Bridge Crack & ZFrac+NN                    & \textbf{99}        & \textbf{97}                   & \textbf{689}         &      \textbf{0.003}                  \\
             & VGG19             & 97         & 96                & 6,902                          & 0.360        \\
             & InceptionV3               & 97      & 95                  & 8,431                &     0.270                  \\
             & ResNet152V2                   & 98       & 93                  & 4,544                 & 0.315               \\
             & DenseNet102                 & 97         & 96                & 3,822                 & 0.341              \\
             \hline
Magnetic  Defect & ZFrac+NN                    & \textbf{99}          & \textbf{98}                 &    \textbf{85}       & \textbf{0.004}                         \\
             & VGG19             & 94            & 94             & 1,644                            & 0.316      \\
             & InceptionV3               & 95       & 94                 & 4,852                    &     0.273              \\
             & ResNet152V2                   & 93      & 92                   & 5,320                 &     0.349           \\
             & DenseNet102                 & 95        & 93                  & 1,024                  &  0.325           \\
             \hline
Anomaly Detection & ZFrac+NN                    & \textbf{94}      & \textbf{92}                     &    \textbf{206}   & \textbf{0.249}                           \\
             & VGG19             & 89            & 85             & 2,434                 & 0.291                 \\
             & InceptionV3               & 91        & 87                & 4,818                &     0.271                  \\
             & ResNet152V2                   & 89       & 85                  & 3,572                &     0.305           \\
             & DenseNet102                 & 89           & 86               & 3,749                 &     0.321         \\
             \hline

Kolektor SDD2 & ZFrac+NN    & 95 & \textbf{96} &   \textbf{891} & \textbf{0.004} \\     
             & VGG19       & 95 & 95 & 1,305 & 0.301              \\
             & InceptionV3   &  \textbf{96} & 93 &   3,481  &0.281           \\
             & ResNet152V2  &   \textbf{96}   & 95  & 3,207 &0.304         \\
             & DenseNet102    &   95   & 94 &    1,984 &   0.325     \\
             \hline
Electroluminescence & ZFrac+NN    & \textbf{85} & \textbf{87} & \textbf{864} & \textbf{0.003}                 \\
 Defective             & VGG19    & 81 &  78 & 1,634 & 0.298                    \\
  Solar Cells           & InceptionV3  &  85 & 79 & 3,045 & 0.282              \\
             & ResNet152V2        & 85 & 79 & 3,011 & 0.306            \\
             & DenseNet102   & 85 & 80 & 3,297 & 0.361                  \\
             \hline
 \hline
             
\end{tabular}
\caption{Performance of ZFrac+NN in comparison to DL models and SOTA models in terms of accuracy, F1 score (\%), training time, and average inference time (sec). The best results are highlighted in bold.}\label{tbl:performance}
\end{table*}

\subsubsection{Robustness}
When considering anomaly or defect detection tasks, it is unclear how machine learning models would generalize to data where anomalies manifest themselves in less significant differences from the training data manifold.

In what follows, we investigate if a fractal feature provides better generalization guarantees on new data since the nature of the fractal feature reflects the object's self-structure. For this purpose, we consider a new dataset related to disaster detection as in \cite{mouzannardamage}. In a digital image, a disaster can be defined as a massive perturbance in the structure of objects which could be well studied in the fractal geometry framework. To test the robustness of our approach, we train ZFrac+NN and 4 DL models that are initialized by random weights - for a fair comparison, on images labeled as \textit{damage} and \textit{non-damage}. The \textit{damage} class has diverse categories ranging from earthquakes to floods and hurricanes. During testing, we consider images showing damage as \textit{fire}, which was never encountered during training.  Table~\ref{tbl:robustness} shows that ZFrac+NN not only achieves better performance on small datasets but it generalizes well on data distributions not encountered during training. This can be explained by the fact that the fractal features are reflecting the sub-structures in the image, which can directly relate to chaos and disaster detection. ZFrac+NN learns how to detect defects faster and more robustly. 
\begin{table}[h]
	\centering
	\begin{tabular}{p{4cm}rrr}
		Number of training images &200 & 400 & 800\\
		\hline
		\hline
		ZFrac+NN  & \textbf{65} & \textbf{71} & \textbf{75} \\
		VGG19 & 42 & 43 & 47 \\
		InceptionV3 & 44 & 46 & 59 \\
		ResNet152V2 & 47 & 49 & 61 \\
		DenseNet102 & 51 & 53 & 61 \\
		
		\hline
	\end{tabular}
\caption{Accuracy (\%) when testing on \textit{fire category}}\label{tbl:robustness}
\end{table}

\subsubsection{Complexity} 
\begin{figure}[h]
    \centering
    \begin{subfigure}[b]{0.5\textwidth}
            \includegraphics[width=0.95\textwidth]{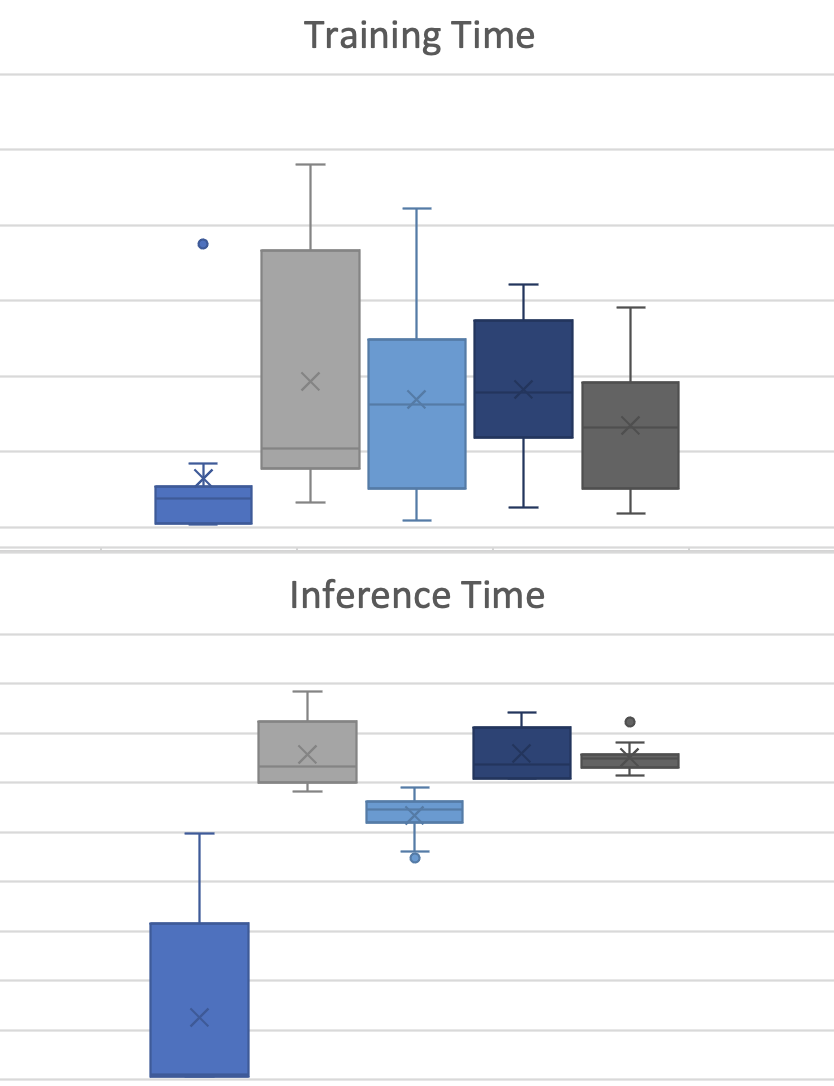}
    \end{subfigure}
        \begin{subfigure}[b]{0.5\textwidth}
            \includegraphics[width=0.95\textwidth]{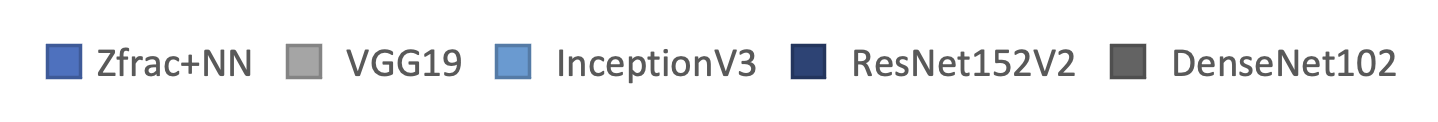}
    \end{subfigure}
    \caption{Training and inference time of ZFrac+NN and the four DL models}\label{fig:time_whisk}
\end{figure}

Studying the computational cost of the extraction of fractal features has a great impact on the efficiency of our method. One could argue that the computation grows exponentially with the number of space coordinates. However, when dealing with $M\times M$ images, computing FD of $(\frac{M}{w})^2$ sub-images of size $w\times w$ takes a number of FLOPS of order $(\frac{M}{w})^2$. Thus, for $\log M$ windows, the number of FLOPs is of order $M^5\log M$, which is computationally cheap given that $M$ is less than $1000$ in most of the applications. 
One can also note that FD computation in a sub-image is independent of others, thus it is easily parallelizable, which can further speed up the training and inference process. Finally, processing the fractal features by a shallow network requires $<10$ MB of memory storage whereas DL models require $>300$MB which might not be always feasible on older generation hardware or for some real-life applications with limited resources such as IoT. Figure~\ref{fig:time_whisk} shows the training and inference times of Table~\ref{tbl:main} aggregated across datasets shows that training ZFrac+NN is more consistent than that of DL models where there is a low variance in the training time. When it comes to the deployed models, the inference time for ZFrac+NN is consistently less than that of the DL models however with higher variance. This can be explained by the fact that ZFrac+NN does not require an image resize. Given that the sizes of the images in the datasets considered in this work vary from 224x224 to 1,024x1,024, the time needed to extract the fractal feature vector will vary considerably. DL models which operate on 224x224 or 229x229 images mostly require a consistent inference time. For applications where a fast inference is crucial, images can be always resized for deployed models.

\subsection{Limitations}
FD aims at describing the self-structure of objects which makes it more suitable for binary classification of defect detection in particular. General classification such as the work on CIFAR and ImageNet data requires more than self-substructure such as colors, context knowledge, and object abstraction. For instance, ZFrac+NN achieves a 67, 35 and 15\% accuracy on MNIST, PETA, and CIFAR, datasets \cite{tang2019improving,byerly2020branching,kolesnikov2019big} while the SOTA performance is 99, 80, and 99\% respectively. This can be explained by the nature of the datasets and the great efforts that have been put into optimizing the performance of DL models. For instance, the PETA data is of low resolution and was taken at a far distance. In such cases, the magnification process in the fractals computation fails to identify substructures, which leads to a weak image understanding, thus poor results. Moreover, the classification of CIFAR and MNIST data requires complex and high-level features that are not included in ZFrac+NN. 

One can thus conclude that the fractal feature has great potential in classification tasks where the self-structure is the essence of that task such as the case of defect detection. However, considered solely, the fractal feature has limitations in classification tasks where higher cognitive knowledge is required. 

\section{Conclusion}\label{sec:conc}
In this work, we inspected deep networks' performance within the fractal geometry framework and we showed that DL models are unable to extract fractal features across different depth levels. We empirically demonstrate that shallow networks trained on the fractal feature exhibit performance comparable to that of DL models that are trained on raw images while requiring less training time in specific tasks. In addition to that, given that fractal geometry studies the internal structure of complex objects, we reveal that such features can lead to better and more robust results in tasks where the study of the object substructure is crucial for classification. We also provide a theoretical complexity analysis showing that computing the fractal feature is efficient. We finally augment our study with a human evaluation of the wrong predictions by DL models as well as ZFrac+NN. The human evaluation showed that, on average, classifications missed by the fractal features are perceived as ``not obvious'' and harder to detect than those of DL models. A subsequent step in this line of work can focus on utilizing the fractal geometry in DL through architecture adaptation.   

\bibliography{cits}


\end{document}